\documentclass[letterpaper, 10 pt, journal, twoside]{IEEEtran}
\usepackage{hyperref}

\usepackage{graphics}           
\usepackage{times}              
\usepackage{amsmath}            
\usepackage{amssymb}            
\usepackage{graphicx}
\usepackage{algorithm}
\usepackage[noend]{algpseudocode}
\usepackage{booktabs}
\usepackage{color}
\definecolor{instructioncolor}{rgb}{.5,.5,.5}

\usepackage[font=small]{caption}

\def\secref#1{Sec.~\ref{#1}}
\def\figref#1{Fig.~\ref{#1}}
\def\tabref#1{Tab.~\ref{#1}}
\def\eqref#1{Eq.~(\ref{#1})}

\makeatletter
\usepackage{xspace}
\DeclareRobustCommand\onedot{\futurelet\@let@token\@onedot}
\def\@onedot{\ifx\@let@token.\else.\null\fi\xspace}
\def\eg{e.g\onedot} 
\def\ie{i.e\onedot}

\def\etal{{et al}\onedot}
\makeatother

\def\etalcite#1{\etal~\cite{#1}}

\usepackage{array}
\newcolumntype{L}[1]{>{\raggedright\let\newline\\\arraybackslash\hspace{0pt}}m{#1}}
\newcolumntype{C}[1]{>{\centering\let\newline\\\arraybackslash\hspace{0pt}}m{#1}}
\newcolumntype{R}[1]{>{\raggedleft\let\newline\\\arraybackslash\hspace{0pt}}m{#1}}

\renewcommand{\b}[1]{\mbox{\boldmath$#1$}}

\renewcommand{\v}[1]{{\b #1}}

\usepackage{pifont}
\newcommand{\xmark}{\ding{55}}%

\usepackage{transparent}
\usepackage{graphicx}
\usepackage{multirow}
\usepackage{amsmath}
\graphicspath{{pics/}}
\usepackage{bm}
\usepackage[normalem]{ulem}

\usepackage[acronyms, shortcuts]{glossaries}
\glsdisablehyper

\newacronym{cnn}{CNN}{convolutional neural network}
\newacronym{rnn}{CNN}{recurrent neural network}
\newacronym{mos}{MOS}{moving object segmentation}
\newacronym{bev}{BEV}{bird's eye view}
\newacronym{knn}{kNN}{k-nearest neighbor}
\newacronym{fifo}{FIFO}{first in, first out}
\newacronym{iou}{IoU}{intersection-over-union}

\newcommand{\miou}{$\text{IoU}_{\text{MOS}}$\xspace}
\newcommand{\mioup}{\mbox{\miou$\left[\%\right]$}}
\newcommand{\filter}{binary Bayes filter\xspace}
\newcommand{\Filter}{Binary Bayes Filter\xspace}
\newcommand{\strategy}{receding horizon strategy\xspace}
\newcommand{\Strategy}{Receding Horizon Strategy\xspace}
\newcommand{\Baseline}{Non-overlapping Strategy\xspace}
\newcommand{\baseline}{non-overlapping strategy\xspace}
\newcommand{\scores}{moving object confidence scores\xspace}
\newcommand{\score}{moving object confidence score\xspace}

\title{Receding Moving Object Segmentation \\ in 3D LiDAR Data Using Sparse 4D Convolutions}

\author{Benedikt Mersch, Xieyuanli Chen, Ignacio Vizzo, Lucas Nunes, Jens Behley, Cyrill Stachniss%
\thanks{Manuscript received: February 24, 2022; Revised: May 10, 2022; Accepted: June 07, 2022. This paper was recommended for publication by Editor Markus Vincze upon evaluation of the Associate Editor and Reviewers' comments.}%
\thanks{All authors are with the University of Bonn, Germany. Cyrill Stachniss is additionally with the Department of Engineering Science at the University of Oxford, UK.}%
\thanks{This work has partially been funded by the Deutsche Forschungsgemeinschaft (DFG, German Research Foundation) under Germany's Excellence Strategy, EXC-2070 -- 390732324 -- PhenoRob and by the European Union’s Horizon 2020 research and innovation programme under grant agreement No~101017008~(Harmony).
}%
\thanks{Digital Object Identifier (DOI): see top of this page.}
}

\begin{document}
\maketitle

\markboth{IEEE Robotics and Automation Letters. Preprint Version. Accepted June, 2022}
{Mersch \MakeLowercase{\textit{et al.}}: Receding Moving Object Segmentation in 3D LiDAR Data Using Sparse 4D Convolutions}

\begin{abstract}
A key challenge for autonomous vehicles is to navigate in unseen dynamic environments. Separating moving objects from static ones is essential for navigation, pose estimation, and understanding how other traffic participants are likely to move in the near future. In this work, we tackle the problem of distinguishing 3D LiDAR points that belong to currently moving objects, like walking pedestrians or driving cars, from points that are obtained from non-moving objects, like walls but also parked cars. Our approach takes a sequence of observed LiDAR scans and turns them into a voxelized sparse~4D point cloud. We apply computationally efficient sparse~4D convolutions to jointly extract spatial and temporal features and predict \scores for all points in the sequence. We develop a \strategy that allows us to predict moving objects online and to refine predictions on the go based on new observations. We use a \filter to recursively integrate new predictions of a scan resulting in more robust estimation. We evaluate our approach on the SemanticKITTI \acl{mos} challenge and show more accurate predictions than existing methods. Since our approach only operates on the geometric information of point clouds over time, it generalizes well to new, unseen environments, which we evaluate on the Apollo dataset.
\end{abstract}

\section{Introduction}
\label{sec:intro}
\IEEEPARstart{D}{istinguishing} moving from static objects in 3D LiDAR data is a crucial task for autonomous systems and required for planning collision-free trajectories and navigating safely in dynamic environments. \Ac{mos} can improve localization~\cite{chen2021ral,chen2019iros}, planning~\cite{thomas2021icra}, mapping~\cite{chen2021ral}, scene flow estimation~\cite{baur2021iccv,gojcic2021cvpr,tishchenko2020threedv}, or the prediction of future states~\cite{toyungyernsub2021icra,wu2020cvpr}. There are mapping approaches that identify if observed points are \emph{potentially moving} or \emph{have moved} throughout the mapping process~\cite{arora2021ecmr,chen2019iros,hornung2013ar,pomerleau2014icra}. On the contrary, identifying objects that are \emph{actually moving} within a short time horizon are of interest for online navigation~\cite{thomas2021icra}, can improve scene flow estimation between two consecutive point clouds~\cite{baur2021iccv,gojcic2021cvpr,tishchenko2020threedv}, or support predicting a future state of the environment~\cite{wu2020cvpr}.

In this work, we focus on the task of segmenting moving objects online using a limited time horizon of observations. Given a sequence of 3D LiDAR scans, we predict for each point if it belongs to a moving object, for example bicyclists or driving cars, or a static, \ie, non-moving one, like parked cars, buildings, or trees.

\begin{figure}[t]
	\centering
	\vspace{0.3cm}
	\def\svgwidth{0.99\linewidth}
	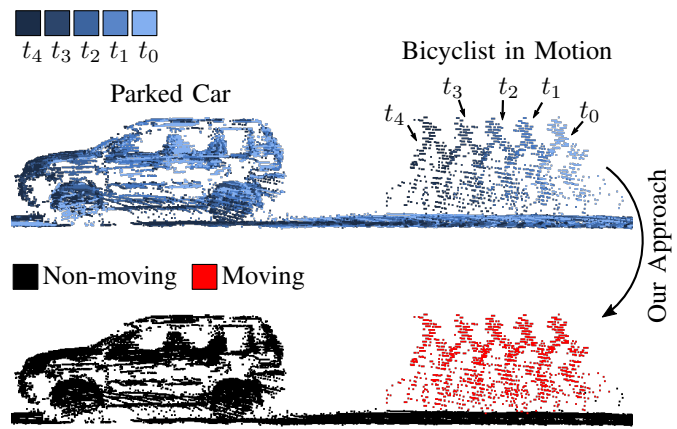
	\caption{Given a sequence of point clouds, our method identifies that the points belonging to the bicyclist move in space over time. \textit{Top}: Input sequence with points colorized (blue) with respect to the time step. The darker the blue color, the newer the scan. \textit{Bottom}: Our method successfully predicts the bicyclist as moving (red) and the parked car as static (black).}
	\label{fig:motivation}
\end{figure}

In contrast to the task of semantic segmentation, \acl{mos} in 3D LiDAR data does not require a complex notion of semantic classes with extensive labeling to supervise learning-based methods or to evaluate their performance. Instead, the goal is to predict if a local point cloud structure moves throughout space and time or remains static as visualized in~\figref{fig:motivation}. In general, the task requires the extraction of temporal information from the LiDAR sequence to decide which points are moving and which are not. Previous works tackled this problem by extracting temporal information from residual range images~\cite{chen2021ral} or \ac{bev} images~\cite{mohapatra2022visapp}, typically using a 2D \ac{cnn}. The back-projection from these 2D representations to the 3D space often requires post-processing like \ac{knn} clustering~\cite{chen2021ral,cortinhal2020arxiv,duerr2020threedv,milioto2019iros} to avoid labels bleeding into points that are close in the image space but distant in 3D\@. Other approaches can identify objects that have moved in 3D space directly during mapping~\cite{arora2021ecmr} or with a clustering and tracking approach~\cite{chen2022ral}. Nevertheless, these offline methods often rely on having access to all LiDAR observations in the sequence.

The main contribution of this paper is a novel approach that predicts moving objects online for a short sequence of LiDAR scans. We exploit sparse~4D convolutions to jointly extract spatio-temporal features from the input point cloud sequence. The outputs of our network are \scores for the points in each input scan. Since we directly predict in a voxelized sparse~4D space, we do not require any back-projection and clustering to retrieve per-point predictions. Our method operates in a sliding window fashion and appends a new observed scan to the input sequence while discarding the oldest one. By doing so, our method can include new observations into the estimation as they arrive. We implemented a \filter to fuse these predictions and in this way increase the robustness to false predictions. Since our method uses only the spatial point information over time, it is class agnostic and generalizes well on unseen data.

In sum, we make three key claims:
Our approach 
(i) segments moving objects in LiDAR data more accurately compared to existing methods,
(ii) generalizes well to unseen environments without additional domain adaptation techniques, and
(iii) improves the results by integrating online new observations. We back explicitly up these three claims by our experimental evaluation. The code of this paper as well as our pre-trained models will be available at \mbox{\url{https://github.com/PRBonn/4DMOS}}.

\begin{figure*}[t]
	\centering
	\fontsize{8}{8}\selectfont
	\def\svgwidth{0.99\linewidth}
	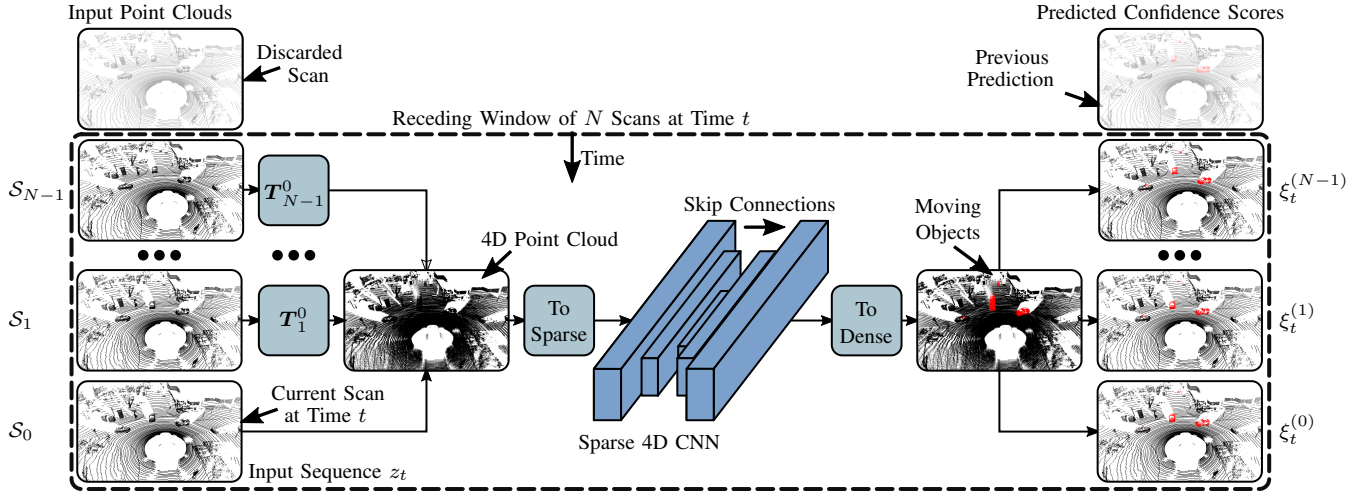
	\caption{Overview of our approach operating with a \strategy. We transform all scans of the considered receding window to the current viewpoint. Next, we aggregate all points and create a sparse~4D point cloud. We apply sparse~4D convolutions to jointly extract spatio-temporal features. Our final layer predicts \scores for all points in the input sequence.}
	\label{fig:approach}
\end{figure*}
\section{Related Work}
\label{sec:related}
We can group LiDAR-based \acl{mos} methods with respect to their definition of dynamic objects. Besides map cleaning methods~\cite{kim2020iros,pomerleau2014icra}, there are also mapping approaches that remove objects that have moved throughout the mapping process from the data before fusing them with the map~\cite{arora2021ecmr,chen2019iros,hornung2013ar}. For example, Wang~\etalcite{wang2012icra} apply graph-based clustering to segment objects that could move in 3D LiDAR data. Ruchti and Burgard~\cite{ruchti2018icra} use a deep neural network to predict dynamic probabilities for each point in a range image before fusing them with a map. In contrast to that, Thomas~\etalcite{thomas2021icra} proposed a self-supervised method for classifying indoor LiDAR points into dynamic labels. The authors explicitly distinguish between short-term and long-term movable objects to treat them differently in localization and planning. Arora~\etalcite{arora2021ecmr} explore ground segmentation with ray-casting to coarsely remove dynamic objects in LiDAR scans. Other researchers encode non-static objects into the map by estimating multi-modal states~\cite{stachniss2005aaai}. Recently, Chen~\etalcite{chen2022ral} propose a pipeline to automatically label moving objects offline. They first use an occupancy-based method to find dynamic point candidates and further identify moving objects by sequential clustering and tracking. Instead of removing all long-term changes caused by objects that have moved, our method segments motion online and focuses on objects that are actually moving within a limited time horizon. 

Previously, scene-flow methods first classify moving points and then estimate separate flows for static and moving objects between two point clouds~\cite{baur2021iccv,gojcic2021cvpr,tishchenko2020threedv}. In more detail, Baur~\etalcite{baur2021iccv} estimate the 3D scene flow between two point clouds composed of a rigid body motion for static and a per-point flow for moving objects. They use a self-supervised motion segmentation signal based on the discrepancy between per-point flow and rigid body motion for training their network. Even though \acl{mos} can be a by-product of scene flow estimation, most methods only consider two subsequent frames which could be a too short time horizon for classifying slowly moving objects.

Other methods primarily focus on segmenting moving objects online using a larger time horizon. To cope with the computational effort of 3D point cloud sequences, projection-based methods have been proposed. Chen~\etalcite{chen2021ral} developed LMNet, which exploits existing single-scan semantic segmentation networks that get residual range images as additional inputs to extract temporal information. Recently, Mohapatra~\etalcite{mohapatra2022visapp} introduced a method using \ac{bev} images for \acl{mos} and achieve faster runtime but inferior performance compared to LMNet. Projection-based methods often suffer from information loss or back-projection artifacts and require additional steps like \ac{knn} clustering~\cite{chen2021ral,cortinhal2020arxiv,duerr2020threedv,milioto2019iros}. In contrast, our method directly predicts in~4D space and does not require any post-processing techniques.

Extracting temporal information from sequential point cloud data is gaining more attention in research since it allows to increase temporal consistency for classification tasks or to predict future states of the environment. To fuse independent semantic single-scan predictions, Dewan and Burgard~\cite{dewan2020icra} use a binary Bayes filter by propagating previous predictions to the next scan using scene flow. In contrast, Duerr~\etalcite{duerr2020threedv} optimize a recurrent neural network to temporally align range image features from a single-scan semantic segmentation network. Some works project the spatial information into 2D representations like range images~\cite{chen2021ral,duerr2020threedv,laddha2020arxiv,mersch2021corl} or \ac{bev} images~\cite{luo2018cvpr,mohapatra2022visapp,wu2020cvpr} and then apply 2D or 3D convolutions to reduce the computational burden of jointly processing~4D data. Besides point-based methods~\cite{fan2021iclr,fan2019arxiv,liu2019iccv} for processing point cloud sequences, representing point clouds as sparse tensors can also circumvent the back-projection issue and makes it possible to apply sparse convolutions efficiently. For example, Shi~\etalcite{shi2020cvpr} propose SpSequenceNet for~4D semantic segmentation which processes two LiDAR frames with sparse 3D convolutions and combines their temporal information with a cross-frame global attention module. To apply convolutions across time, Choy~\etalcite{choy2019cvpr} propose Minkowski networks for semantic segmentation using sparse~4D convolutions on temporal RGB-D data.

In this paper, we propose a novel \acl{mos} method that jointly applies sparse~4D convolutions on a sequence of LiDAR point clouds building on top of the Minkowski engine~\cite{choy2019cvpr}. Unlike previous methods, we operate online and do not need a pre-built map representation. Whereas most classification methods output one prediction for each frame, we propose to predict moving objects using a \strategy. This allows us to incorporate new observations in an online fashion and refine predictions more robustly by Bayesian filtering.

\section{Our Approach}\label{sec:main}
Given a point cloud sequence~\mbox{$\mathcal{S}{=}\{\mathcal{S}_j\}_{j=0}^{N-1}$}of~$N$~LiDAR scans~\mbox{$\mathcal{S}_{j}{=}\{\v{p}_{i} {\in} \mathbb{R}^4\}_{i=0}^{M_j-1}$} with~$M_j$ points represented as homogeneous coordinates, \ie,~$\v{p}_i {=} \left[x_i,y_i,z_i,1\right]^\top$, the goal of our approach is to predict, which points are actually moving in the input sequence~$\mathcal{S}$. We denote the current scan as~$\mathcal{S}_0$ and index the previous past scans from~$1$ to~$N{-}1$.

As shown in~\figref{fig:approach}, we first transform the past point clouds~$\mathcal{S}_1,\dots,\mathcal{S}_{N-1}$ to the viewpoint of the current scan~$\mathcal{S}_0$ and create a sparse~4D tensor, see~\secref{sec:representation}. We extract spatio-temporal features with a sparse convolutional architecture and predict confidence scores of being actually moving for each point in the sequence, see~\secref{sec:network}. As soon as we obtain a new LiDAR scan, we shift the prediction window as explained in~\secref{sec:receding_prediction}. The \strategy allows to recursively update the estimation by fusing later predictions for the same scan in the sequence in a \filter, see~\secref{sec:fusion}.

\subsection{Input Representation}\label{sec:representation}
The first step is to locally align all past point clouds~$\mathcal{S}_1,\dots,\mathcal{S}_{N-1}$ in the sequence~$\mathcal{S}$ to the viewpoint of the current LiDAR scan~$\mathcal{S}_0$. In this work, we assume to have access to \emph{estimated} relative pose transformations~$\v{T}_j^{j-1}$ between scans~$\mathcal{S}_{j-1}$ and~$\mathcal{S}_j$. Odometry estimation is a standard task for autonomous vehicles and can be efficiently solved on-board with an online SLAM system like SuMa~\cite{behley2018rss} and further improved by integrating information from an inertial measurement unit~\cite{shan2020iros} or by using wheel encoders. Our approach is agnostic to the source of odometry information and a local consistency is sufficient such that obtaining this data is not a problem in practice. We represent the relative transformations between scans~$\v{T}_{1}^{0}, \dots, \v{T}_{N}^{N-1}$ as transformation matrices, \ie,~$\v{T}_j^{j-1}{\in}\mathbb{R}^{4\times4}$. Further, we denote the~$j^\text{th}$ scan transformed to the current viewpoint by
\begin{align}
\mathcal{S}^{j \rightarrow 0} & =\{\v{T}_j^0\v{p}_{i}\}_{\v{p}_i \in \mathcal{S}_j} \quad
\text{with}~\v{T}_j^0 = \prod_{k = 0}^{j-1} \v{T}_{j-k}^{j-k-1}.
\label{eq:transformed_scan}
\end{align}

The motivation behind locally aligning the scans in the sequence is that our \ac{cnn} should focus on local point patterns that move in space over time and for that, pose information helps. We also provide an experimental analysis on the effect of the pose alignment in~\secref{sec:ablation}. After applying the transformations, we aggregate the aligned scans into a~4D point cloud by converting from homogeneous coordinates to cartesian coordinates and by adding the time as an additional dimension resulting in coordinates~$\left[x_i,y_i,z_i,t_i\right]^\top$~for point~$\v{p}_i$.

Since outdoor point clouds obtained from a LiDAR sensor are sparse by nature, we quantize the~4D point cloud into a sparse voxel grid with a fixed resolution in time~$\Delta t$ and space~$\Delta s$. We use a sparse tensor to represent the voxel grid and store the indices and associated features of non-empty voxels only. Sparse tensors are more memory efficient compared to dense voxel grids since they only store information about the voxels that are actually occupied by points. The sparse representation allows us to use spatio-temporal \acp{cnn} efficiently since common dense~4D convolutions become intractable on large scenes.

\subsection{Sparse 4D Convolutions}\label{sec:network}
Using the sparse input representation discussed in~\secref{sec:representation}, we can apply time- and memory-efficient sparse~4D convolutions to jointly extract spatio-temporal features from the sparse~4D occupancy grid and predict a \score for each point. To this end, we use the Minkowski engine~\cite{choy2019cvpr} for sparse convolutions. Sparse convolutions operate on the sparse tensor and define kernel maps that specify how the kernel weights connect the input and output coordinates. The main advantage of sparse convolutions is the computational speed-up compared to dense convolutions.

We use a sparse convolutional network developed for~4D semantic segmentation on RGB-D data and adapt it for \acl{mos} on LiDAR data. More specifically, we use a modified MinkUNet14~\cite{choy2019cvpr}, which is a sparse equivalent of a residual bottleneck architecture with strided sparse convolutions for downsampling the feature maps and strided sparse transpose convolutions for upsampling. The skip connections in a UNet fashion~\cite{ronneberger2015micc} help to maintain details and fine-grained predictions. We reduce the number of feature channels in the network resulting in a model with~$1.8$\,M parameters, which is comparably low compared to the \acl{mos} baseline LMNet~\cite{chen2021ral} with SalsaNext~\cite{cortinhal2020arxiv}~($6.7$\,M) or RangeNet++~\cite{milioto2019iros}~($50$\,M) backbones. The last layer of our network is a~4D sparse convolution with a softmax that predicts \scores between~$0$ and~$1$ for each point. 

In contrast to~4D semantic segmentation methods that use RGB values as input features~\cite{choy2019cvpr}, we initialize voxels occupied by at least one point with a constant feature of~$0.5$. Therefore, our input is a sparse~4D occupancy grid only storing voxels occupied by a point. This makes it easier to deploy the approach in new environments without estimating the distribution of coordinates or intensity values to standardize the input data as done for semantic segmentation~\cite{milioto2019iros}. The generalization capability of our approach is further investigated in~\secref{sec:generalization}. 

\subsection{\Strategy}\label{sec:receding_prediction}
The fully sparse convolutional architecture introduced in~\secref{sec:network} jointly predicts \scores for all points in the input sequence. At inference time, one option would be to divide the input data into fixed, non-overlapping intervals and to predict each sub-sequence once. 

Instead, we propose a different strategy and develop a \strategy for \acl{mos}. When the LiDAR sensor obtains the next point cloud, we add it to the input sequence and discard the oldest scan resulting in a \acl{fifo} queue, see~\figref{fig:approach}. The main advantage is that we can re-estimate moving objects based on new observations and therefore increase the time horizon used for prediction. It is a natural idea to use multiple observations to reduce the uncertainty of semantic estimations and has been well investigated in mapping algorithms like SuMa++~\cite{chen2019iros}. It is still rarely used for online segmentation, and we propose a method to improve the online \acl{mos}.

\subsection{\Filter}\label{sec:fusion}
\begin{figure}[t]
	\centering
	\fontsize{8}{8}\selectfont
	\def\svgwidth{0.99\linewidth}
	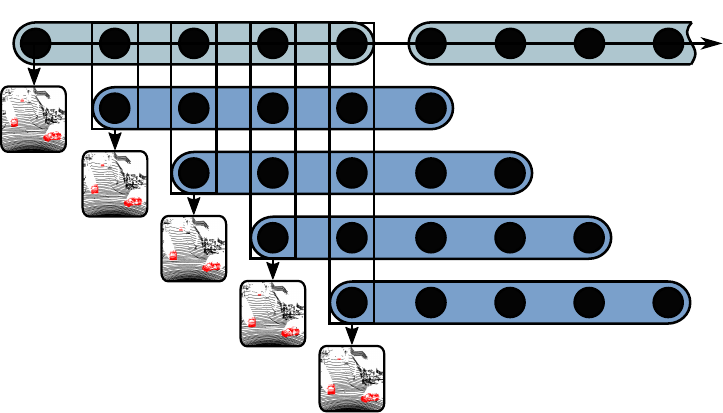
	\caption{Overview of our proposed \filter. At~$t{=}5$, the \baseline uses the five per-scan confidence predictions, whereas our \strategy integrates the next observation at~$t{=}6$ by shifting the temporal window. Our \filter then fuses multiple \scores to improve the prediction.}
	\label{fig:bayesian_fusion}
	\vspace{-0.3cm}
\end{figure}

Since our proposed method predicts moving objects in~$N$~scans at once, the \strategy leads to a re-estimation of the previously predicted~$N{-}1$~scans. These multiple predictions from different time steps allow refining the estimation of moving objects based on new observations. We propose to fuse them recursively using a \filter. This makes it possible to increase the time horizon used for segmentation and helps to predict slowly moving objects that only moved a small distance within the initial time horizon. The Bayesian fusion reduces the number of false positives and negatives that arise due to occlusions or noisy measurements.

More formally, for a scan~$\mathcal{S}_j$, we can estimate moving objects at time~$t$ by fusing all predicted \scores from previously observed point cloud sequences~$z_{0:t}$ that contain the scan~$\mathcal{S}_j$. The term~$z_t$ denotes the observed input point cloud sequence~$\mathcal{S}_0,\dots,\mathcal{S}_{N-1}$ with $\mathcal{S}_0$ recorded at time $t$. We want to estimate the joint probability distribution of the moving state~$m^{(j)}$ of all points up to time~$t$ denoted by
\begin{equation}\label{eqn:joint_prob}
p\left(m^{(j)} \mid z^{(j)}_{0:t}\right) = \prod_{i} p\left(m^{(j)}_i \mid z^{(j)}_{0:t}\right),
\end{equation}
where~$m^{(j)}_i{\in}\{ 0,1 \}$ is the state of point~\mbox{$\v{p}_{i} {\in} \mathcal{S}_j$} being moving in the scan~$\mathcal{S}_j$. For better readability, we will from now on consider a single point~$\v{p}_{i}$ in point cloud~$\mathcal{S}_j$ and omit the superscript~$j$ without loss of generality. 

We apply Bayes' rule to the per-point probability distribution~$p\left(m_i \mid z_{0:t}\right)$ in \eqref{eqn:joint_prob} and follow the standard derivation of the recursive binary Bayes filter~\cite{thrun2005probrobbook}. Using the log-odds notation~$l(x){=}\log \frac{p(x)}{1-p(x)}$ commonly used in occupancy grid mapping, we finally end up with
~
\begin{equation}\label{eq:log_odds}
\hspace{0.2cm}
l\left(m_i 
\hspace{-0.1cm}\mid\hspace{-0.1cm} z_{0:t}\right) \hspace{-0.1cm}=\hspace{-0.1cm}
\begin{cases}
l\left(m_i 
\hspace{-0.1cm}\mid\hspace{-0.1cm}
z_{0:t-1}\right)
\hspace{-0.05cm}+\hspace{-0.05cm}
l\left(m_i \
\hspace{-0.1cm}\mid\hspace{-0.1cm}
z_t\right) 
\hspace{-0.05cm}-\hspace{-0.05cm}
l(m_i) , 
&\hspace{-0.2cm} \text{if }t {\in} \mathcal{T}\\
l\left(m_i 
\hspace{-0.1cm}\mid\hspace{-0.1cm} z_{0:t-1}\right) , 
&\hspace{-0.2cm}\text{otherwise},
\end{cases}
\end{equation}
with~$\mathcal{T}$ being the set of time steps in which we observe point~$\v{p}_{i}$ in the input sequence~$z_t$. Whereas~$l\left(m_i \mid z_{0:t-1}\right)$ is a recursive term including all predictions for the point~$i$ up to time~$t{-}1$, the term~$l\left(m_i \mid z_t\right)$ denotes the log-odds of the probability to be moving at time~$t$. Note that if we do not observe the point~$\v{p}_{i}$ at time~$t$, there is no prediction and we do not update the recursive term~$l\left(m_i \mid z_{0:t-1}\right)$. The prior probability~$p_0{\in}(0,1)$ in the last part~$l(m_i){=}\log \frac{p_0}{1-p_0}$ provides a measure of the innovation introduced by a new prediction. For \acl{mos}, the prior determines how much a predicted moving point in a single scan influences the final prediction. We will investigate different priors in~\secref{sec:ablation}.

At time~$t$, our network outputs \scores~\mbox{$\xi_{t}^{\left( j \right)} {\in}  \{ \xi_{t,i} \}_{i=0}^{M_j-1}$} with~$\xi_{t,i} {\in} (0,1)$ for each point cloud~$\mathcal{S}_j$ with~$M_j$ points in the current input sequence~$z_t$. We can interpret the predicted confidence score~$\xi_{t,i}$ for a single point~$\v{p}_{i}$ given the input sequence~$z_t$ as posterior probability reading
\begin{equation}\label{eqn:output}
\xi_{t,i}=p\left(m_i=1 \mid z_t\right).
\end{equation}

The log-odds expression of the confidence score in~\eqref{eq:log_odds} is then given as
\begin{equation}\label{eqn:inverse_sensor_model}
l\left(m_i \mid z_t\right)=
\log \frac{\xi_{t,i}}{1-\xi_{t,i}}.
\end{equation}

\figref{fig:bayesian_fusion} illustrates the \baseline in the upper part and our proposed \strategy with a \filter to fuse multiple predictions in the lower part. We obtain the final prediction by converting the recursively estimated per-point log-odds~$l\left(m_i \mid z_{0:t-1}\right)$ to confidence score using \mbox{$p(x){=}\log \frac{l(x)}{1+l(x)}$}. If the confidence is larger than~$0.5$, we predict the point to be moving and otherwise non-moving.

\section{Experimental Evaluation}
\label{sec:exp}
The main focus of this work is a method to segment actually moving objects in 3D LiDAR data by exploiting consecutive scans in an online fashion. Additionally, we carry out the prediction using a \strategy and integrate new predictions recursively in a \filter.

We present our experiments to show the capabilities of our method and to
support our three key claims: Our approach 
(i) segments moving objects in LiDAR data more accurately compared to existing methods,
(ii) generalizes well to unseen environments without additional domain adaptation techniques, and
(iii) improves the results by integrating online new observations.

\subsection{Experimental Setup}\label{sec:setup}
For our experimental evaluation, we train all models on the SemanticKITTI~\cite{behley2019iccv} dataset. We use sequences $00$-$07$ and $09$-$10$ for training, $08$ for validation, and $11$-$21$ for testing. During training, we optimize the model with a binary cross-entropy loss for all points in the input sequence and a learning rate of~$0.0001$ and a weight decay of~$0.0001$ with the Adam optimizer~\cite{kingma2015iclr}. If not stated differently in the experiments, our input point clouds sequences contain~$N{=}10$ input scans with a temporal resolution of~$\Delta t{=}0.1$\,s. The spatial voxel size for quantization is~$\Delta s{=}0.1$\,m. To increase the diversity of the training data and to avoid overfitting, we follow the data augmentation of Nunes~\etalcite{nunes2022ral} and apply random rotations, shifting, flipping, jittering, and scaling to all points in the same~4D point cloud. We train all networks for less than~$60$ epochs and keep the model with the best performance on the validation set. We follow the \strategy presented in~\secref{sec:receding_prediction} and combine predictions with the \filter proposed in~\secref{sec:fusion} using a prior of~$p_0{=}0.25$.

For quantitative evaluation, we report the standard \ac{iou} metric~\cite{everingham2010ijcv} for the moving class given by
\begin{align}
\text{IoU}_{\text{MOS}} & = \frac{\text{TP}}{\text{TP} + \text{FP} + \text{FN}}, \label{eq:miou}
\end{align}
with true positive~TP, false positive~FP, and false negative~FN classifications of moving points. 

To evaluate the generalization capability of our approach across environments, we additionally test it on another dataset without the use of domain adaptation techniques. We follow the setup of Chen~\etalcite{chen2022ral} and use the Apollo-ColumbiaParkMapData~\cite{lu2019cvpr} dataset sequence~$2$ (frames~$22300$-$24300$) and sequence~$3$ (frames~$3100$-$3600$) annotated the same way as SemanticKITTI\@. Note that SemanticKITTI and Apollo both use Velodyne HDL-64E LiDAR scanners, but they are mounted on a different car at a different height and recorded data in a different environment.

\subsection{Moving Object Segmentation Performance}\label{sec:performance}
Our first experiment evaluates the performance of our model on the SemanticKITTI~\cite{behley2019iccv} \acl{mos} benchmark~\cite{chen2021ral}. The results support the first claim about segmenting moving objects more accurately compared to existing methods that are published and open-source. For a fair comparison, we follow the setup from LMNet~\cite{chen2021ral} and use the provided SemanticKITTI poses estimated with an online SLAM system~\cite{behley2018rss}. We report the result on the hidden test set in~\tabref{tab:benchmark} and compare it to additional baselines provided by Chen~\etalcite{chen2021ral}. 

One can see that single-scan segmentation with SalsaNext~\cite{cortinhal2020arxiv} and predicting all movable classes as moving leads to low performance of~$4.4\%$ \miou. The same applies to estimating scene flow and thresholding the flow vectors to determine if an object moves. The online multi-scan semantic segmentation methods SpSequenceNet~\cite{shi2020cvpr}, LMNet~\cite{chen2021ral}, and KPConv~\cite{thomas2019iccv} show improved results up to~$60.9\%$ \miou, see~\tabref{tab:benchmark}. Our method can outperform all baselines with an \miou of~$65.2\%$, which demonstrates the effectiveness of our approach. Our performance is also better than LMNet+AutoMOS+Extra~\cite{chen2022ral}, which additionally uses automatically generated moving object labels for training. This emphasizes the strength of our result.

\begin{table}[t]
	\centering
	\begin{tabular}{L{6cm}c}
		\toprule
		& \mioup\\
		\midrule
		SalsaNext~\cite{cortinhal2020arxiv} (movable classes)		   		& $4.4$  \\
		SceneFlow~\cite{liu2019cvpr}     				   				& $4.8$  \\
		SpSequenceNet~\cite{shi2020cvpr}             		   				& $43.2$ \\
		LMNet~\cite{chen2021ral}					   					& $58.3$ \\
		KPConv~\cite{thomas2019iccv}               		       			& $60.9$ \\
		Ours, $N{=}10$ Scans, $\Delta t{=}0.1$\,s, $p_0{=}0.25$ & $\mathbf{65.2}$\\
		\midrule
		LMNet+AutoMOS+Extra~\cite{chen2022ral}                  					& $62.3$ \\
		\bottomrule
	\end{tabular}
	\caption{Performance on SemanticKITTI~\cite{behley2019iccv} \acl{mos} benchmark~\cite{chen2021ral}. Baseline results taken from~\cite{chen2022ral}. Best result in bold.}
	\label{tab:benchmark}
\end{table}

\subsection{Generalization Capabilities}\label{sec:generalization}
The next experiment evaluates our method's ability to generalize across different environments. It supports our second claim that the approach generalizes well on unseen data. We test our model on the Apollo dataset without using any domain adaptation techniques or re-training and compare to baselines that use different levels of domain adaptation. LMNet~\cite{chen2021ral} uses the same SemanticKITTI~\cite{behley2019iccv} sequences for training, whereas LMNet+AutoMOS~\cite{chen2022ral} is LMNet trained on an automatically labeled training set of Apollo. LMNet+AutoMOS+Fine-Tuned~\cite{chen2022ral} is a model pre-trained on SemanticKITTI and fine-tuned on Apollo, see~\cite{chen2022ral} for details. The results in~\tabref{tab:apollo} suggest that for the baselines, domain adaptation like re-training or fine-tuning improves the results with a maximum \miou of~$65.9\%$. Our method yields the highest \miou of~$73.1\%$ without any additional steps, which shows that the approach is well capable of predicting moving objects in an unknown environment. 

We hypothesize that extracting moving object features in a sparse~4D occupancy grid is advantageous since the method does not use any sensor-specific information like intensity or RGB values. Directly operating in~4D space also makes the network less prone to overfitting to a specific sensor location as in the case of range-images, where moving objects are usually found in certain areas of the image. We also do not use information about semantic classes whose distribution can differ between environments.

\begin{table}[t]
	\centering
	\begin{tabular}{L{6cm}c}
		\toprule
		& \mioup\\
		\midrule
		LMNet~\cite{chen2021ral}	   							& $16.9$ \\
		LMNet+AutoMOS~\cite{chen2022ral}            				& $45.7$ \\
		LMNet+AutoMOS+Fine-Tuned~\cite{chen2022ral}         		& $65.9$ \\
		Ours, $N{=}10$ Scans, $\Delta t{=}0.1$\,s, $p_0{=}0.25$	& $\mathbf{73.1}$ \\
		\bottomrule
	\end{tabular}
	\caption{Performance on Apollo~\cite{lu2019cvpr} dataset. Best result in bold.}
	\label{tab:apollo}
\end{table}

\subsection{\Strategy and Fusion}\label{sec:prior}
This section backs up our third claim that the proposed \strategy combined with a \filter improves the \ac{mos} results by integrating online new observations. We investigate the effect of using different numbers of input scans~$N$ and temporal resolutions~$\Delta t$ for prediction as well as fusing with different prior probabilities~$p_0$ in the Bayesian fusion presented in~\secref{sec:fusion}. 

\begin{figure}[t]
	\centering
	\includegraphics[width=0.95\linewidth]{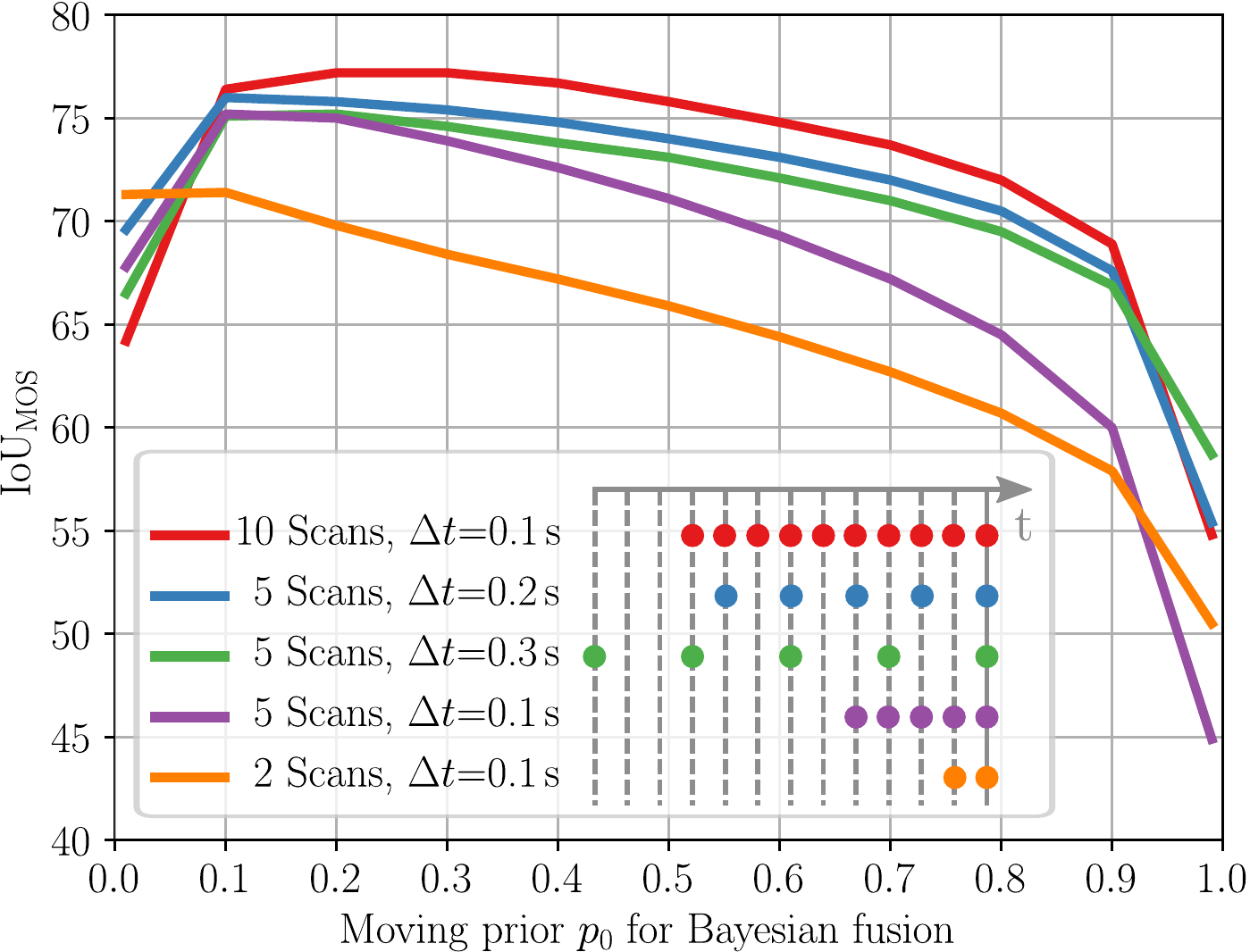}
	\caption{Comparison of \miou on the SemanticKITTI~\cite{behley2019iccv} validation set using different moving object priors for the \filter. The colors indicate variants of our approach using different time horizons and resolutions. The colored dots on the timeline visualize which past scans we input to the model for prediction at time $t$}.
	\label{fig:prior}
\end{figure}

We compare models trained on~$N{=}2$,~$5$, and~$10$ input and output scans as well as a model that predicts a single output scan. Since the combination of a \strategy and the Bayesian fusion of multiple beliefs allows us to use information from a larger time horizon, we additionally compare to two variant setups using~$N{=}5$ input and output frames but with a different temporal resolution. One uses a resolution of~$\Delta t{=}0.2$\,s resulting in a time horizon of~$0.8$\,s, the other one processes~$1.2$\,s of scans that are~$\Delta t{=}0.3$\,s apart. For comparison, the method using~$N{=}2$ scans with a resolution of~$\Delta t{=}0.1$\,s has a time horizon of~$0.2$\,s, the one with~$N{=}5$ scans a horizon of~$0.4$\,s and the model using~$N{=}10$ scans looks at~$0.9$\,s of data. We visualize the time horizons and temporal resolutions for each variant in~\figref{fig:prior} as colored dots on a timeline sampled at~$10$\,Hz.

The Bayesian prior~$p_0$ in~\eqref{eq:log_odds} serves to compute the difference between the new predicted log-odds and the initially expected log-odds. Therefore, modifying the prior influences the contribution of new observed moving objects to the updated prediction. ~\figref{fig:prior} shows the \miou on the SemanticKITTI validation set for different priors. With a small prior (\eg~$p_0{=}0.01$), we fuse predicted moving objects more aggressively leading to more true positives, but also an increased number of false positives since inconsistent predictions are not filtered out. A large prior (\eg~$p_0{=}0.99$) results in a conservative fusion where objects are only predicted to be moving if all predictions agree. We found that a moving object prior between~$0.1$ and~$0.3$ works best for the SemanticKITTI validation sequence. We experienced that for a lot of slowly moving objects in the scene, setting a lower prior helps to keep them in the final prediction even if they have not been predicted moving from all available instances in time. We achieve the best result with a model using~$N{=}10$ input and output frames and a Bayesian prior of~$p_0{=}0.25$, which is the setup for the experiments presented in~\secref{sec:performance} and~\secref{sec:generalization}. 

In general, a combination of processing more scans and fusing multiple predictions with the \strategy works best for achieving a larger time horizon resulting in better \acl{mos}. Our approach also works with fewer scans but with a larger temporal resolution, which reduces the computational effort. The models using~$N{=}5$ input scans with a larger temporal resolution of~$\Delta t{=}0.2$\,s and~$\Delta t{=}0.3$\,s between scans outperform the model with the same number of processed scans but a smaller resolution of~$\Delta t{=}0.1$\,s, which is the actual sensor frame rate. This shows that the extended time horizon achieved with a larger temporal resolution leads to better segmented slowly moving objects since their motion is more visible in the sequence.

\subsection{Ablation Study}\label{sec:ablation}
To further support our third claim and show the effectiveness of individual proposed components of our approach, we train different variants of our network and evaluate their performance on the validation set. We train all models for up to~$60$ epochs and report the best \miou on the validation set during training, see~\tabref{tab:ablation}. For all methods, we compare two prediction strategies: First, a \baseline that divides the input sequence into sub-sequences and predicts each sub-sequence independently, see the upper part in~\figref{fig:bayesian_fusion}. Second, our \strategy proposed in~\secref{sec:receding_prediction}, which generates multiple predictions for the same scan and fuses them in a \filter (again using a prior of~$p_0{=}0.25$). We visualize this combination in the lower part of~\figref{fig:bayesian_fusion}.

In general, we see an improvement of up to~$5.4$ percentage points of \miou for all models using the proposed \strategy. More precisely, using the \filter with model~[A] reduces the number of false negatives by~$8.2$\% and the number of false positives by~$18.9$\%. This indicates that the proposed approach successfully integrates more observations into the estimation and is more robust to false predictions due to occlusions or noisy measurements. If we compare the performance of model~[A] using~$N{=}5$ scans which are~$\Delta t{=}0.1$\,s apart to the networks trained with larger temporal resolutions of~$\Delta t{=}0.2$\,s~[B] and~$\Delta t{=}0.3$\,s~[C], we again see that the results can be further improved by considering a larger time horizon, see also~\secref{sec:prior}. If the point clouds are not transformed into a common viewpoint, the method~[D] is still able to infer moving objects but at a reduced performance of~\miou${=}39.9\%$ with Bayesian fusion. This is because the network needs to infer both the ego-motion of the sensor as well as the relative motion of the objects. When only training to predict a single output scan~[E], the result is worse and fusing more predictions is not possible since no additional predictions are available. Next, one can see that our method can also achieve \acl{mos} with two scans only~[F] but the performance is worse. The best performing model~[G] takes~$N{=}10$ input scans with a temporal resolution of~$\Delta t{=}0.1$\,s and fuses the predictions resulting in an \miou of~$77.2\%$. The results show that we achieve a better \acl{mos} by increasing the time horizon with a combination of processing more scans, increasing the temporal resolution, and using the proposed \strategy with a \filter. 

\begin{table}[t]
	\centering
	\begin{center}
		\begin{tabular}{ccccccc}
			\toprule
			&\# Inputs&\# Outputs&Poses?&$\Delta t$&\multicolumn{2}{c}{\mioup}\\
			\cmidrule{6-7}
			&&&&&w/o BF&w/ BF\\
			\midrule
			{[A]}&$5$&$5$&\checkmark	&$0.1$\,s&$69.1$&$\textbf{74.5}$\\
			{[B]}&$5$&$5$&\checkmark	&$0.2$\,s&$71.8$&$\textbf{75.6}$\\
			{[C]}&$5$&$5$&\checkmark	&$0.3$\,s&$71.6$&$\textbf{74.9}$\\
			{[D]}&$5$&$5$&\xmark		&$0.1$\,s&$35.6$&$\textbf{39.9}$\\
			{[E]}&$5$&$1$&\checkmark	&$0.1$\,s&$\textbf{66.5}$&-\\
			{[F]}&$2$&$2$&\checkmark	&$0.1$\,s&$64.9$&$\textbf{69.0}$\\
			{[G]}&$10$&$10$&\checkmark&$0.1$\,s&$74.3$&$\textbf{77.2}$\\
			\bottomrule
		\end{tabular}
		\caption{Ablation study on different variants of our approach with and without the proposed \strategy and Bayesian fusion (BF) using a prior~$p_0{=}0.25$. We denote the temporal resolution between scans by~$\Delta t$. Best results in bold.}
		\label{tab:ablation}
	\end{center}
\end{table}

\subsection{Qualitative Results}\label{sec:qualitative}
\begin{figure}[t]
	\centering
	\fontsize{8}{8}\selectfont
	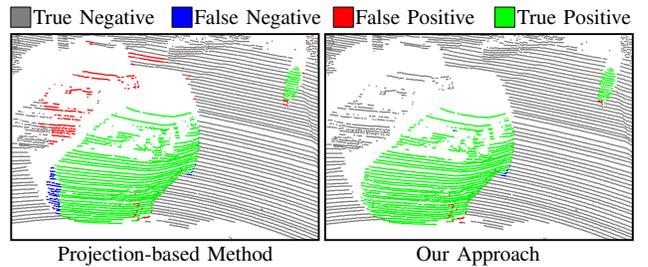
	\caption{Qualitative comparison of segmentation accuracy. \textit{Left}:~Prediction by range image-based LMNet~\cite{chen2021ral} after \ac{knn} post-processing. \textit{Right}:~Our sparse voxel-based approach without further post-processing. Best viewed in color.}
	\label{fig:qualitative1}
\end{figure}

Finally, we illustrate that our method predicts actually moving objects in 3D space without the need for geometric post-processing like clustering.  We use the model from~\secref{sec:performance} trained on~$N{=}10$ scans with a temporal resolution of~$\Delta t{=}0.1$\,s. In~\figref{fig:qualitative1}, we show the segmentation of scan~$1638$ from the SemanticKITTI~\cite{behley2019iccv} validation sequence~$08$. We compare the range image-based method LMNet~\cite{chen2021ral} to our sparse voxel-based approach. One can see that despite the geometric-based \ac{knn} post-processing, the baseline still shows artifacts and bleeding labels behind the moving car illustrated as red-colored false positives whereas our method directly predicts in the 3D space without boundary effects.

\begin{figure}[t]
	\centering
	\fontsize{8}{8}\selectfont
	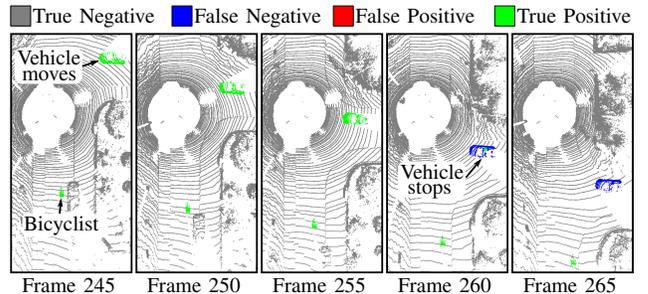
	\caption{Change in \acl{mos} if an object stops moving. Best viewed in color.}
	\label{fig:qualitative2}
\end{figure}

Next, \figref{fig:qualitative2} shows how the prediction changes for a scene in the validation set in which a vehicle stops moving. Since our method does not use any semantic understanding of objects, it only reasons about how the points move in space for the given time horizon. Since the vehicle stops moving to yield at the intersection, our method's prediction changes from moving to static. The bicyclist in the back is successfully classified as moving. Note that this results in a false negative indicated in blue since the ground truth SemanticKITTI labels consider if an object has moved throughout the data collection and not based on recent movement.


\subsection{Runtime}\label{sec:runtime}
With our unoptimized Python implementation, the network requires on average~$0.078$\,s for predicting moving objects in~$10$ scans and~$0.047$\,s for~$5$ scans both using an~NVIDIA~RTX~A5000. Our \filter only adds a small overhead of~$0.008$\,s on average for fusing~$10$ predictions and~$0.004$\,s for fusing~$5$ predictions.

\section{Conclusion}
\label{sec:conclusion}
In this paper, we present a novel approach to segment moving objects in 3D LiDAR data. Our method jointly predicts moving objects for all scans in the input sequence and operates using a \strategy. We report improved performance on the SemanticKITTI \acl{mos} benchmark and show that the approach generalizes well on unseen data. Our proposed \strategy in combination with a \filter allows us to extend the time horizon used for segmenting moving objects and to increase the robustness to false positive and false negative predictions. Currently, we estimate odometry and \acl{mos} separately which can be jointly optimized in future work.

\bibliographystyle{plain_abbrv}

\bibliography{references}

\end{document}